# Bernoulli Embeddings for Graphs

Vinith Misra[α] and Sumit Bhatia[β*]
[α]Netflix Inc., Los Gatos, CA, USA
[β]IBM India Research Laboratory, New Delhi, India
vmisra@netflix.com, sumitbhatia@in.ibm.com

March 25, 2018

## Abstract

Just as semantic hashing [Salakhutdinov and Hinton2009] can accelerate information retrieval, binary valued embeddings can significantly reduce latency in the retrieval of graphical data. We introduce a simple but effective model for learning such binary vectors for nodes in a graph. By imagining the embeddings as independent coin flips of varying bias, continuous optimization techniques can be applied to the approximate expected loss. Embeddings optimized in this fashion consistently outperform the quantization of both spectral graph embeddings and various learned real-valued embeddings, on both ranking and pre-ranking tasks for a variety of datasets.

## 1 Introduction

Consider users — perhaps from the research, intelligence, or recruiting community — who seek to explore graphical data — perhaps knowledge graphs or social networks. If the graph is small, it is reasonable for these users to directly explore the data by examining nodes and traversing edges. For larger graphs, or for graphs with noisy edges, it rapidly becomes necessary to algorithmically aid users. The problems that arise in this setting are essentially those of information retrieval and recommendation for graphical data, and are well studied [Hasan and Zaki2011, Blanco et al.2013]: identifying the most important edges, predicting links that do not exist, and the like. The responsiveness of these retrieval systems is critical [Gray and Boehm-Davis2000], and has driven numerous system designs in both hardware [Hong et al.2011] and software [Low et al.2014]. An alternative is to seek *algorithmic* solutions to this latency challenge.

Models that perform link prediction and node retrieval can be evaluated across two axes: the relevance of the retrieved nodes and the speed of retrieval. The gold standard in relevance is typically set by trained models that rely on "observable" features that quantify the connectivity between two nodes, but these models are often quite slow to evaluate due to the complexity of the features in question.

At the other extreme, binary-valued embeddings can accelerate the retrieval of graphical data, much in the same manner that semantic hashing [Salakhutdinov and Hinton2009] can assist in the efficient retrieval of text and image data. Roughly speaking, having a binary representation for each node allows one to search for similar nodes in constant time directly in the binary embedding space — much faster than the alternatives (Table 1).

The challenge is that efficient binary representations can be difficult to learn: for any reasonable metric of accuracy, finding optimal binary representations is NP-hard. One solution is to lean on the large body of work around learning *continuous* embeddings for graphs, and utilize modern quantization techniques to binarize these continuous representations. The "catch" with this approach is that the continuous embeddings are not optimized with their future binarization in mind, and this hurts the relevance of retrieved nodes (Sec. 4.4).

---

[*]Part of this work was conducted while both the authors were at IBM Almaden Research Center.

| Technique | Preprocess | Query |
|---|---|---|
| Observable features | $O(1)$ (slow) | $O(N)$ (slow) |
| Real embeddings | $O(ED)$ | $O(N)$ (fast) |
| **Binary embeddings (sparse similarities)** | $O(ED)$ | $O(1)$ |
| Real embeddings (quantized) | $O(ED)$ | $O(1)$ |
| Binary embeddings (dense similarities) | $O(N^2)$ | $O(1)$ |

Table 1: Complexity of five different node retrieval approaches, ranked from highest to lowest accuracy (Table 2). $N$ nodes, $E$ edges, and $D$ latent dimensions.

Our primary contribution, thus, is an *end-to-end method for learning embeddings that are explicitly optimized with both binarization and their use in link prediction/node retrieval in mind*. More concretely: in a manner similar to Skip-gram [Mikolov *et al.*2013], the likelihood of an edge between two nodes is modeled as a function of the Hamming distance between their bit embeddings. Rather than directly optimizing the bit embeddings $e_{ij}$ for this task — an NP-hard task [Weiss *et al.*2009] — they are instead imagined as being drawn from a matrix of independent Bernoulli random variables $E_{ij}$, parametrized by their (independent) probabilities of success $p_{ij}$. By minimizing *expected* loss over this (product) distribution of embeddings, and by applying efficient approximations to the Hamming distance (Sec. 3.4), continuous optimization techniques can be applied. For convenience, we refer to bit embeddings learned in this manner as *Bernoulli embeddings*.

Comparisons performed on five different graphical datasets are described in Section 4. Bernoulli embeddings are found to achieve significantly higher test-set mean average precision than a variety of alternative binary embedding options, including various quantizations of DeepWalk vectors [Perozzi *et al.*2014], Fiedler embeddings [Hendrickson2007], and several other real-valued embeddings that we ourselves introduce (Table 2). This is also found to hold for the reranking scenario, where binary hashes are used as a preprocessing step to accelerate more computationally intensive algorithms. Further, node retrieval performed using binary embeddings is orders of magnitude faster than other alternatives, especially for larger datasets (Table 4).

## 2 Related Work

Approaches to node retrieval, roughly categorized in Table 1, can be evaluated in terms of both relevance and speed. Bernoulli embeddings occupy an unexplored but valuable niche in this spectrum: they are binary embeddings ($O(1)$ retrieval) appropriate for large graphs (more than a few thousand nodes) that are learned directly from the adjacency matrix (higher relevance of retrieved nodes). In the following, we describe the other categories represented in Table 1.

### 2.1 Observable features

Methods for link prediction and node retrieval on graphs typically rely on observable neighborhood features [Dong *et al.*2014]. However, computing node-node similarity using these tools can have a significant computational cost [Low *et al.*2014].

Second order neighborhood features, such as the Jaccard index [Hasan and Zaki2011] and the Adamic-Adar (AA) score [Adamic and Adar2003] either implicitly or explicitly involve operations over length-2 paths and degree-2 neighborhoods. This generally requires either one or more joins with the graph table or multiplications with the graph's adjacency matrix. Even with efficient sparsity-exploiting indexing, this operation has complexity $O(E)$ in the number of edges $E$ in the graph.

Higher order path features, such as the Katz metric, rooted PageRank [Hasan and Zaki2011], and the regression-based Path Ranking Algorithm [Dong *et al.*2014] involve even longer paths and even more such oper-ations. State-of-the-art link prediction often harnesses dozens of such features in parallel [Cukierski *et al.*2011]. Offline precomputation of the (dense) node similarity matrix can dramatically help with latency, but its quadratic complexity in the number of nodes leaves it an option only for smaller graphs.

## 2.2 Real-valued Embeddings

With unquantized real-valued embeddings (i.e. no use of LSH), node retrieval typically involves a brute-force $O(N)$ search for the nearest Euclidean neighbors to a query. While such embeddings have appeared most prominently in the context of text for reasons unrelated to retrieval, Perozzi et al [Perozzi *et al.*2014] apply the word2vec machinery of Mikolov et al [Mikolov *et al.*2013] to "sentences" generated by random walks on a graph, and Yang et al [Yang *et al.*2015] illustrate that a graph embedding can be a powerful tool in the context of semisupervised learning. The world of knowledge graphs has been particularly welcoming to embeddings, starting with the work of Hinton [Hinton1986], and continuing with the models of Bordes et al. [Bordes *et al.*2011], Sutskever et al. [Sutskever *et al.*2009], Socher et al. [Socher *et al.*2013] and others.

Additionally, graph Laplacian eigenvectors, which find prominent use in spectral clustering, can be interpreted as a latent embedding ("Fiedler embedding" [Hendrickson2007]) analogous to LSA. Knowledge graphs, which are more naturally represented with tensors than with adjacency matrices, analogously suggest the use of tensor factorizations and approximate factorizations [Nickel *et al.*2012].

## 2.3 Discrete Embeddings

Hinton and Salakhudtinov [Salakhutdinov and Hinton2009] introduce semantic hashing as the solution to a very similar problem in a different domain. Instead of relying on indexed TF-IDF vectors for the retrieval of relevant documents, a discrete embedding is learned for every document in the corpus. At query time, a user can rapidly retrieve a shortlist of relevant documents simply by scanning the query's neighborhood in the (discrete) embedding space. If the embedding is sufficiently compact, the neighborhood will be nonempty and the scan will be fast. If the embedding is *very* compact, this retrieval yields a pre-ranked list that may be reranked using more computationally demanding algorithms. These results have fueled the development of a variety of similar techniques, all seeking to learn compact binary encodings for a given dataset.

**Quantized Real-valued Embeddings:** The most popular approach, taken by Weiss et al. [Weiss *et al.*2009], Gong and Lazebnik [Gong and Lazebnik2011] and others, is to assume that the data consists of short real vectors. To apply these algorithms to graphical data, one must first learn a real-valued embedding for the nodes of the graph. We compare against these baselines in Sec. 4.

**Binary Embeddings from Similarity Matrix:** In this approach, also known as "supervised hashing"' [Liu *et al.*2012, Kulis and Grauman2012, Norouzi *et al.*2012], a matrix of pairwise similarities between all data points is supplied. The objective is to preserve these similarities in the discrete embedding space. On the surface, this appears very similar to the graphical setting of interest to us. However, *no sparsity assumption* is placed on the similarity matrix. As such, proposed solutions (typically, variations of coordinate descent) fall victim to an $O(N^2)$ complexity, and application is limited to graphs with a few thousand nodes. Note that an embedding-learning approach specifically avoids this issue by exploiting the sparse structure of most graphs.

Liu et al. [Liu *et al.*2014] also assume that one is supplied a matrix of similarities for the data. Rather than directly attempting to replicate these similarities in the embedded space, they perform a constrained optimization that forces the embedded bits to be uncorrelated and zero-mean. In the graph setting, this acts as an approximation to the sign bits of the Fiedler embedding, which appears amongst our empirical baselines.

# 3 Architecture

## 3.1 Bernoulli Embeddings

We consider a generic graph, consisting of $N$ nodes $\mathcal{X} \triangleq \{1, 2, \ldots, N\}$ and a binary-valued adjacency matrix $G \in \{0, 1\}^{N \times N}$.

The goal is formulated as learning a matrix of probabilities $\mathbf{p} = \{p_{ij} : 1 \leq i \leq N, \, 1 \leq j \leq d\}$, from which the node embeddings $E$ are sampled as a matrix of independent Bernoulli random variables $E_{ij} \sim \beta(p_{ij})$. For convenience, one may reparameterize the embeddings as $E_{ij} = \mathbb{1}(p_{ij} > \Theta_{ij})$, where $\Theta^{N \times d}$ is a matrix of iid random thresholds distributed uniformly over the interval $[0, 1]$.

## 3.2 Model and Objective

Recall the use case for short binary codes: to retrieve a shortlist of nodes $Y \in \mathcal{X}^m$ similar to a query node $x \in \mathcal{X}$, one should merely have to look up entries indexed at nearby locations in the embedding space. As such, we seek a model where *the Hamming distance between embeddings monotonically reflects the likelihood of a connection between the nodes.*

For real-valued embeddings, a natural and simple choice — used for instance with Skip-gram [Mikolov *et al.*2013] — is to treat the conditional link probability between nodes $i$ and $j$ as a softmax-normalized cosine distance between embeddings:

$$\mathbf{P}(j|i; \mathbb{p}) = \frac{\exp(E_i \cdot E_j)}{\sum_{k=1}^{|\mathcal{X}|} \exp(E_i \cdot E_k)} = \text{softmax}_j \left[ E_i^T E \right]. \tag{1}$$

To translate this to the setting of binary vectors, it is natural to substitute Hamming distance $d_H(E_i, E_j)$ for the cosine distance $E_i \cdot E_j$ in (1). As the distance $d_H$ is limited to taking values in the set $\{0, 1, \ldots, d\}$, a transformation is required. Empirically, we find that more complex transformations are unnecessary for the purpose of learning a good embedding[1], and a simple linear scaling suffices

$$\mathbf{P}(j|i; \mathbf{p}) = \text{softmax}_j \left[ a d_H(E_i^T, E) \right], \tag{2}$$

where $d_H(x, y) = x^T(1-y) + (1-x)^T y$ represents the Hamming distance, and $a$ is the (potentially negative) scaling parameter.

Given the model of (2), we seek to maximize the expected log likelihood of the observed graph edges:

$$L(G; \mathbf{p}) = \sum_{(i,j) \in G} -\mathbf{E} \left[ \log \text{softmax}_j \left( a E_i^T E \right) \right]_\Theta. \tag{3}$$

The expression in (3) unfortunately introduces two obstacles: (1) the softmax, which requires a summation over all candidate nodes $j$, and (2) the expectation of a discrete-valued functional, which presents difficulties for optimization. The first is addressed by means of noise contrastive estimation [Gutmann and Hyvärinen2010], detailed in Sec. 3.3. The second is addressed via several approximation techniques, detailed in Sec. 3.4.

## 3.3 Noise Contrastive Estimation (NCE)

To sidestep the softmax summation, we follow in the steps of Mnih and Kavukcuoglu [Mnih and Kavukcuoglu2013] and employ NCE [Gutmann and Hyvärinen2010], whose minimum coincides with that of (3). Specifically, softmax normalization is replaced with a learnable parameter $b$, and one instead optimizes for the model's effectiveness at distinguishing a true data point $(i,j) \in G$ from randomly generated noise $(i, K_{ij})$. This objective is given by

$$L(G) = \sum_{(i,j)\in G} -\mathbf{E}\left[\log \frac{e^{ad_H(E_i,E_j)+b}}{e^{ad_H(E_i,E_j)+b} + p_K(j|i)} + \log \frac{p_K(K_{ij}|i)}{e^{ad_H(E_i,E_{K_{ij}})+b} + p_K(K_{ij}|i)}\right]_\Theta, \quad (4)$$

where $K_{ij}$ is a negative sample drawn from the conditional noise distribution $p_K(\cdot|i)$.

---

[1] More specifically: while a complex choice of transformation can improve the optimization objective (4), we find no consistent or significant improvement to the test set precision-recall metrics reported in Sec. 4.4. Essentially, beyond a point, parametrization of the mapping appears to improve the probabilities produced in a forward pass through the network, but does not noticeably improve the embedding-parameter-gradients it returns in the backwards pass.

Gutmann and Hyvärinen [Gutmann and Hyvärinen 2010] argue that one should choose the noise distribution to resemble the data distribution as closely as possible. We experiment with distributions ranging from powers of the unigram, to distributions over a node's 2nd-degree neighborhood (in accordance with the locally closed world assumption of Dong et al. [Dong et al. 2014]), to mixtures thereof, to curricula that transition from easily identified noise to more complex noise models. Empirically, we find that none of these techniques outperform the uniform noise distribution either significantly or consistently.

### 3.4 Approximation of objective

The objective function in (4) presents a challenge to gradient-based optimization. The expectation over $\Theta$ is difficult to evaluate analytically, and because the argument to the expectation is discrete, the reparameterization trick [Kingma and Welling 2013] does not help. We introduce two continuous approximations to the discrete random variable $D_H(E_i, E_j)$ that maneuver around this difficulty.

First, according to the independent-Bernoulli model for the embedding matrix $E_{ij} = \mathbb{1}(p_{ij} > \Theta_{ij})$, the normalized Hamming distance between two embeddings is the mean of $d$ independent (but not identically distributed) Bernoullis $F_1, \ldots, F_d$:

$$\frac{1}{d}D_H(E_i, E_j) = \frac{1}{d}\sum_{l=1}^d E_{il}(1-E_{jl}) + (1-E_{il})E_{jl} = \frac{1}{d}\sum_{l=1}^d F_l.$$

By Kolmogorov's strong law [Sen and Singer 1994], this quantity converges with $d$ almost surely to its expectation. Therefore, for sufficiently large $d$, the $\Theta$-expectation in (4) is approximated by

$$L_{\text{mean}}(G) = \sum_{(i,j)\in G} -\log \frac{e^{ad_H(\mathbf{p}_i,\mathbf{p}_j)+b}}{e^{ad_H(\mathbf{p}_i,\mathbf{p}_j)+b} + p_K(j|i)} - \log \frac{p_K(K_{ij}|i)}{e^{ad_H(\mathbf{p}_i,\mathbf{p}_K)+b} + p_K(K_{ij}|i)}, \quad (5)$$

which is amenable to gradient-based optimization.

While the approximation of (5) is accurate for larger dimensionality $d$, recall that our goal is to learn *short* binary codes. A sharper approximation is possible for smaller $d$ by means of the central limit theorem as follows.

$$\frac{1}{d}D_H(E_i, E_j) \approx \mathcal{N}\left(\mu_{ij} = D_H(\mathbf{p}_i, \mathbf{p}_j), \sigma_{ij}^2 \right.$$
$$\left. = \sum_{k=1}^d D_H(\mathbf{p}_{ik}, \mathbf{p}_{jk})(1 - D_H(\mathbf{p}_{ik}, \mathbf{p}_{jk}))\right). \quad (6)$$

Applying the reparametrization trick [Kingma and Welling 2013], our objective takes the form

$$L_{\text{CLT}}(G) = \sum_{(i,j)\in G} -\mathbf{E}\left[\log \frac{e^{a(\mu_{ij}+\sigma_{ij}Z)+b}}{e^{a(\mu_{ij}+\sigma_{ij}Z)+b} + p_K(j|i)} \right.$$
$$\left. + \log \frac{p_K(K_{ij}|i)}{e^{a(\mu_{ij}+\sigma_{ij}Z)+b} + p_K(K_{ij}|i)}\right]_Z, \quad (7)$$

where $Z$ is a zero mean and unit variance Gaussian. Observe that the argument to the expectation is now differentiable with respect to the parameters $(\mathbf{p}, a, b)$, as is the case with (5).

A common approach [Kingma and Welling 2013] to optimizing an expected-reparameterized loss such as $L_{\text{CLT}}$ is to use Monte Carlo integration to approximate the gradient over $N$ samples of noise $Z$. This yields an unbiased estimate for the gradient with variance that converges $O\left(\frac{1}{N}\right)$. However, Monte Carlo integration is generally appropriate for approximating *higher* dimensional integrals. For a *single* dimensional integral, numerical quadrature — while deterministic and therefore biased — can have the significantly faster error convergence of $O\left(\frac{1}{N^4}\right)$ (midpoint rule). For small $N$, accuracy can be further improved by performing quadrature with respect to the Gaussian CDF $\Phi_z$. Letting $f$ denote the intra-expectation computation in (7),

$$
\begin{aligned}
\mathbf{E}\left[f(\mu_{ij} + \sigma_{ij} Z)\right]_Z &= \int_0^1 f(\mu_{ij} + \sigma_{ij} z) d\Phi_z \\
&\approx \frac{1}{2N} \sum_{n=1}^N f\left(\mu_{ij} + \sigma_{ij} \Phi_z^{-1}\left(\frac{2n-1}{2N}\right)\right).
\end{aligned}
\tag{8}
$$

Figure 1 compares the mean approximation with the quadrature normal approximation over the range of embedding dimensionalities we consider. For smaller embedding dimensionalities, the greater accuracy of the quadrature approximation leads to a lower test set log loss.

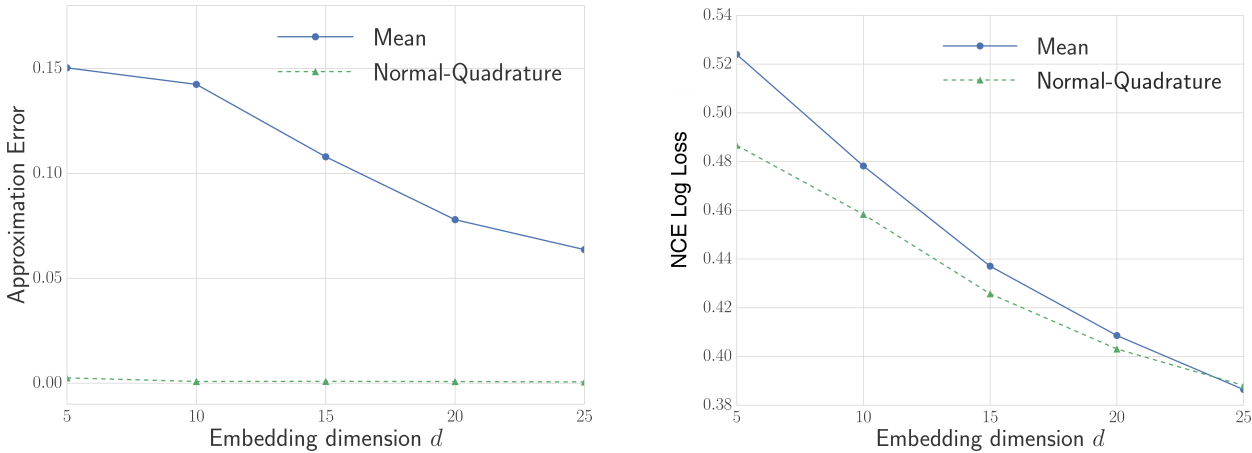

Figure 1: Comparison of Bernoulli models optimized for $L_{\text{mean}}$ and for $L_{\text{CLR}}$ ($N = 5$ samples) on the KG dataset. Left: Error (absolute) between approximate loss and true loss. Right: True loss.

## 3.5 Optimization and Discretization

The training set loss, as given by $L_{\text{CLT}}$ and approximated by (8), is minimized with stochastic gradient descent with the diagonalized AdaGrad update rule [Duchi *et al.* 2011]. To generate a discrete embedding $E = \mathbb{1}(\mathbf{p} < \Theta)$ from a Bernoulli matrix $\mathbf{p}$, each entry is rounded to 0 or 1 (i.e. $\Theta = 1/2$), in accordance with maximum likelihood.

# 4 Experiments

## 4.1 Datasets

Results are evaluated on five datasets. Five percent of the edges of each dataset are held out for the test set, and the remainder are used in training.

**KG** (115K entities, 1.3M directed edges)[2] is a knowledge graph extracted from the Wikipedia corpus using statistical relation extraction software [Castelli *et al.*2012]. Edges are filtered to those with more than one supporting location in the text, and both entity and relation types are ignored. Despite this filtration, KG possesses a large number of spurious edges and entities. Such a noisy dataset is representative of automatically constructed knowledge graphs commonly encountered in enterprise settings [Bhatia *et al.*2016].

**Wordnet** (82K entities, 232K directed edges) is a comparatively low-noise, manually constructed graph consisting of the noun-to-noun relations in the Wordnet dataset [Miller1995]. As with KG, we ignore the edge type attribute.

**Slashdot** (82K entities, 948K directed edges), **Flickr** (81K entities, 5.9M undirected edges), and **Blog-Catalog** (10K entities, 334K undirected edges) are standard social graph datasets consisting of links between users of the respective websites [Leskovec *et al.*2009, Tang and Liu2009].

## 4.2 Baselines and Comparisons

While there is little work specifically in obtaining bit-valued embeddings for graphs, we compare Bernoulli embeddings against quantizations of three classes of real-valued embeddings.

**B1: Fiedler embeddings** [Hendrickson2007] are computed using the unnormalized graph Laplacian, symmetric graph Laplacian, and random-walk graph Laplacian.

**B2: DeepWalk embeddings** are computed using the Skip-gram-inspired model of Perozzi et al [Perozzi *et al.*2014].

**B3: Real-valued distance embeddings (DistEmb)** are obtained by modifying the Bernoulli embedding objective to predict link probabilities from the Hamming, $\ell^1$, $\ell^2$, or cosine distance between real-valued embedded vectors. Note that the Hamming distance case is equivalent to using the Bernoulli probabilities **p** as embeddings, and the cosine distance variety can be interpreted as DeepWalk modified with NCE and a window size of 2.

Three different quantizations of the above embeddings are computed. Random-hyperplane LSH [Charikar2002] is selected due to its explicit goal of representing cosine similarity — used by both the DeepWalk and the cosine-distance variety of DistEmb. Spectral Hashing (SH) is chosen for its similarity in objective to Fiedler embeddings. Iterative Quantization (IQ), another data-driven embedding, has been found to outperform SH on several datasets [Gong and Lazebnik2011], and as such we consider it as well.

**B4: Observable predictor.** Additionally, for use in re-ranking, we perform logistic regression with several observable neighborhood features of the form $s(x,y) = \sum_{z \in \Gamma(x) \cap \Gamma(y)} f(\Gamma(z))$, where $\Gamma(x)$ indicates the degree-1 neighborhood of $x$. Specifically, we compute the number of common neighbors ($f(x) = x$), the Adamic-Adar (AA) score ($f(x) = 1/\log(x)$), variations of AA ($f(x) = x^{-0.5}, x^{-0.3}$), and transformations $T(s) = [s, \log(s+1), s^{0.5}, s^{0.3}, s^2]$ of each score. Despite being far from state-of-the-art, we find that this predictor can significantly improve performance when reranking results produced with binary embeddings.

Both 10- and 25-dimensional embeddings are trained: for the scale of graphs we consider, the (sparsely populated) latter is useful for instant-retrieval via semantic-hashing, while the (densely populated) former is useful for reranking. Furthermore, we find that the quantizations of the real embeddings **B1**-**B3** perform best when highly-informative 100 and 200 dimensional Fiedler/DeepWalk/DistEmb embeddings are quantized down to the 10 and 25 bit vectors that are sought.

## 4.3 Evaluation metrics

Each of the methods considered is likely to excel in the metric it is optimized for: Bernoulli embeddings for expected log loss, DeepWalk for Skip-gram context prediction, etc. Our interest, however, lies in node retrieval, and more specifically in the ranked list of nodes returned by each algorithm. Mean Average Precision (MAP) is a commonly used and relatively neutral criterion appropriate for this task.

---
[2] http://sumitbhatia.net/source/datasets.html

A subtlety, however, lies in the choice of set on which to evaluate MAP. Document retrieval algorithms commonly evaluate precision and recall on documents in the training set [Salakhutdinov and Hinton 2009]. This does not lead to overfitting, as the algorithms typically only make use of the training set documents and not their labeled categories/similarities. Embedding-based link-prediction algorithms, however, *explicitly* learn from the labeled similarity information, as represented by the edge list / adjacency matrix. Alternatively stated: rather than extrapolating similarities from input text, the goal is to generalize and predict *additional* similarities from those that have already been observed.

As such, we measure generalization via "test set precision": for a query node, a retrieved node is only judged as relevant if an edge between it and the query appears in the test set. Observe that this is much smaller than typically reported MAP, as all edges appearing in the training set are judged non-relevant. More specifically, for small test sets, its value can rarely be expected to exceed the inverse of the average degree.

Furthermore, in reporting observed results, scores corresponding to "DistEmb", "DeepWalk", and "Fiedler" embeddings are the *best* observed test set score amongst all variations of quantizer type (SH, LSH, and ITQ), graph laplacian type (unnormalized, symmetric, random walk), and distance embedding (Hamming and $\ell^2$ [3]). These optimizations almost certainly represent test set overfitting, and present a challenging baseline for the Bernoulli embeddings.

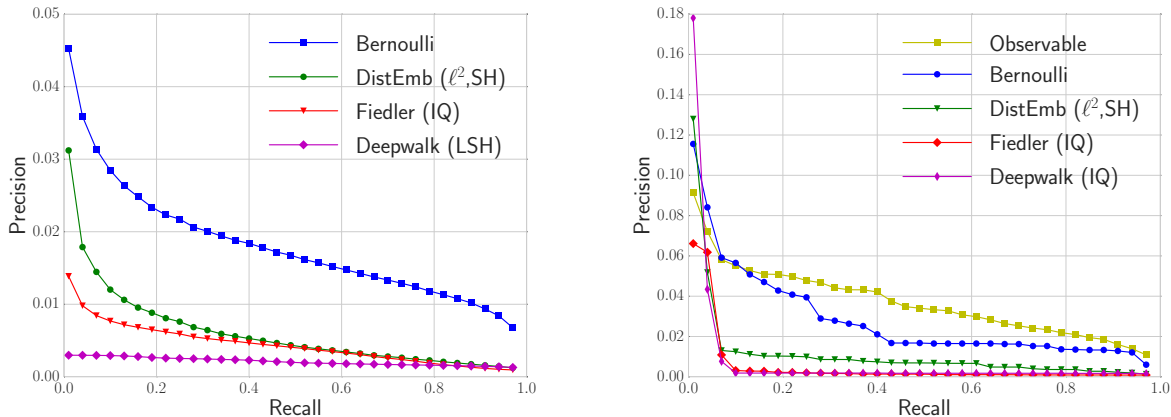

Figure 2: Left: 25-bit Ranking. Right: 10-bit Reranking. Mean precision/recall of binary embeddings on the Flickr test set, averaged over 1000 random queries. Real embeddings are quantized to 25 bits from 100 dimensions, and only their highest test set MAP parameterizations are shown.

### 4.4 Empirical results

In the case of directly retrieving results from binary embeddings, Bernoulli embeddings are found to significantly outperform the various alternative binary embeddings (Fig. 2 and Table 2). It is also interesting to compare to the *unquantized* real embeddings (last four rows of Table 2). Despite their informational disadvantage, Bernoulli embeddings are competitive.

On most datasets, the observable-feature predictor achieves significantly higher MAP than any of the latent embedding models — real or binary. This is in line with our expectations, and one can in fact expect even better such results from a state-of-the-art link predictor. To harness this predictive power without the computational expense of computing observable features, a compelling option is to *re-rank* the highest-confidence nodes retrieved by binary embeddings [Salakhutdinov and Hinton 2009]. Observe the two computational bottlenecks at play here: searching in the embedded space to populate the pre-ranked list of entities, and computing observable features for each of the pre-ranked entities.

---
[3] $\ell^1$ and cos similarity are consistently outperformed.

| Dataset | | KG | | Wordnet | | Slashdot | | Flickr | | BlogCatalog | |
|---|---|---|---|---|---|---|---|---|---|---|---|
| $d$ (Dimensionality) | | 10 | 25 | 10 | 25 | 10 | 25 | 10 | 25 | 10 | 25 |
| Ranked, Binary $E_i \in \{0,1\}^d$ | Q-DistEmb | .0016 | .0047 | .0021 | .0212 | .0014 | .0213 | .0051 | .0054 | .0073 | .0094 |
| | Q-DeepWalk | .0004 | .0004 | .0011 | .0012 | .0002 | .0004 | .0014 | .0017 | .0048 | .0053 |
| | Q-Fiedler | .0011 | .0026 | .0001 | .0004 | .0009 | .0216 | .0021 | .0039 | .0065 | .0080 |
| | Bernoulli | **.0042** | **.0112** | **.0054** | **.1013** | **.0016** | **.0298** | **.0087** | **.0161** | **.0131** | **.0181** |
| Reranked, Binary $E_i \in \{0,1\}^d$ | Q-DistEmb | .0122 | .0179 | .0074 | .0178 | .0471 | **.0493** | .0092 | .0084 | .0098 | .0198 |
| | Q-DeepWalk | .0049 | .0060 | .0031 | .0030 | .0081 | .0189 | .0056 | .0112 | **.0196** | .0210 |
| | Q-Fiedler | .0091 | .0129 | .0019 | .0024 | .0420 | .0402 | .0044 | .0074 | .0139 | .0172 |
| | Bernoulli | **.0249** | **.0267** | **.0101** | **.0514** | **.0516** | .0487 | **.0240** | **.0487** | .0181 | **.0241** |
| Observables | | .0865 | | .0126 | | .0898 | | .0349 | | .0524 | |
| Ranked, Real $E_i \in R^d$ | DistEmb | .0216 | .0254 | .0195 | .1227 | .0313 | .0350 | .0196 | .0256 | .0255 | .0282 |
| | Fiedler | .0085 | .0086 | .0020 | .0016 | .0269 | .0270 | .0065 | .0077 | .0118 | .0116 |
| | DeepWalk | .0022 | .0021 | .0761 | .0777 | .0326 | .0326 | .0037 | .0037 | .0074 | .0074 |

Table 2: Test set MAP at both 10 and 25 dimensions. For DistEmb, DeepWalk, Fiedler, and quantizations Q-* we present the maximum MAP across Laplacian varieties, choice of DistEmb distance function, and choice of quantizer. Bold indicates category leaders.

We re-rank under constraints on both of these operations: no more than 10000 locations in the embedded space may be searched, and no more than 1000 nodes can be subject to re-ranking. As illustrated in Fig. 2 and documented in the second four rows of Table 2, Bernoulli embeddings are again found to outperform the alternatives.

Finally, note that while both the Bernoulli objective function and the test-set MAP evaluation center around the prediction of *unknown* links, generalization with those metrics also translates into a more qualitatively meaningful ranking of *known* links. Table 3 illustrates this for the KG dataset. Notice that the observed quality of the retrieved entities roughly reflects the MAP scores of the respective predictors/embeddings.

| Query | Rank | DistEmb ($\ell^2$,SH) | Bernoulli | Bernoulli Reranked | Observable |
|---|---|---|---|---|---|
| "New Delhi" | 1st | "Ministry of Education" | "Delhi" | "Delhi" | "Delhi" |
| | 2nd | "Indonesian" | "Mumbai" | "India" | "India" |
| | 3rd | "Poet" | "British India" | "Indian" | "Alumni" |
| "Roger Federer" | 1st | "Gael Monfils" | "Grand Slam Final" | "Rafael Nadal" | "Rafael Nadal" |
| | 2nd | "Third Round" | "Maria Sharapova" | "French Open" | "French Open" |
| | 3rd | "Second Round" | "Rafael Nadal" | "Novak Djokovic" | "Wimbledon" |
| "Bruce Wayne" | 1st | "Dick Grayson" | "Joker" | "Batman" | "Batman" |
| | 2nd | "Damian Wayne" | "Batman" | "Robin" | "Robin" |
| | 3rd | "Gotham" | "Arkham Asylum" | "Joker" | "Joker" |

Table 3: Top-3 retrieved nodes (excluding the query node) in the KG dataset for several queries.

## 4.5 Efficiency Results

Training a 25-dimensional embedding on the KG dataset — the largest problem we consider — takes roughly 10s per epoch on a 2.2GHz Intel i7 with 16GB of RAM, and validation loss is typically minimized between 30 and 60 epochs.

Table 4 reports the average time taken by different methods to retrieve the node most similar to a query node. By considering a $d$-dimensional binary embedding as an address/hash in the d-dimensional space, retrieval reduces to a near-instant look up of the hashtable: 0.003 ms for 25-dimension embeddings (Table 4). Note that this retrieval speed is independent of the embedding dimensionality $d$ and the dataset size. At the other extreme, the high-performance observables model for the KG dataset (115K entities) takes 2949 ms and scales linearly with the number of nodes in the dataset. Finally, the "Goldilocks" solution takes 7 ms: a 10-dimensional embedding is used to obtain a shortlist, which is then reranked using the observables model.

| Method | Time (ms) |
|---|---:|
| binary embeddings, hash based retrieval | 0.003 |
| ranking using observables (KG) | 2949 |
| re-ranking using observables (KG) | 7.6 |
| binaryBruteForce (100d, 100K nodes) | 0.7 |
| binaryBruteForce (100d, 1M nodes) | 4.5 |
| binaryBruteForce (100d, 10M nodes) | 91.7 |
| binaryBruteForce (100d, 100M nodes) | 1147.9 |
| realBruteForce (100d, 100K nodes) | 23.5 |
| realBruteForce (100d, 1M nodes) | 181.6 |
| realBruteForce (100d, 10M nodes) | 2967.1 |
| realBruteForce (100d, 100M nodes) | OOM |

Table 4: Time taken in milliseconds by different methods for retrieving similar nodes given a query node. Times reported are averaged over 50 runs, ran on a system running Ubuntu 14.10, with 32GB RAM and 16 Core 2.3 GHz Intel Xeon processor. OOM indicates Out of Memory.

Hash based lookup is not the only way to exploit binary embeddings. If one replaces a brute force nearest neighbor search amongst real valued embeddings with a brute force search amongst binary embeddings, order-of-magnitude memory and speed advantages remain despite the $O(N)$ runtime: binary embeddings take up 32 to 64 times less space than real embeddings, and Hamming distance can be computed much faster than Euclidean distance. Table 4 illustrates this on synthetically generated embeddings for web-scale graphs of up to 100 million nodes.

## 5 Conclusion

We introduce the problem of learning discrete-valued embeddings for graphs, as a graphical analog to semantic hashing. To sidestep the difficulties in optimizing a discrete graph embedding, the problem is reformulated as learning the continuous parameters of a distribution from which a discrete embedding is sampled. Bernoulli embeddings correspond to the simplest such distribution, and we find that they are computable with appropriate approximations and sampling techniques. On a variety of datasets, in addition to being memory and time efficient, Bernoulli embeddings demonstrate significantly better test-set precision and recall than the alternative of hashed real embeddings. This performance gap continues to hold when retrieval with binary embeddings is followed by reranking with a more powerful (and cumbersome) link predictor. In this latter case, precision and recall can rival or exceed that of the link predictor, while performing retrieval in time more or less independent of the data set size.